# Belief Propagation by Message Passing in Junction Trees: Computing Each Message Faster Using GPU Parallelization

**Lu Zheng and Ole Mengshoel**[*]
Electrical and Computer Engineering Department,
Carnegie Mellon University.

**Jike Chong**[†]
Electrical Engineering and Computer
Science Department,
University of California, Berkeley.

## Abstract

Compiling Bayesian networks (BNs) to junction trees and performing belief propagation over them is among the most prominent approaches to computing posteriors in BNs. However, belief propagation over junction tree is known to be computationally intensive in the general case. Its complexity may increase dramatically with the connectivity and state space cardinality of Bayesian network nodes. In this paper, we address this computational challenge using GPU parallelization. We develop data structures and algorithms that extend existing junction tree techniques, and specifically develop a novel approach to computing each belief propagation message in parallel. We implement our approach on an NVIDIA GPU and test it using BNs from several applications. Experimentally, we study how junction tree parameters affect parallelization opportunities and hence the performance of our algorithm. We achieve speedups ranging from 0.68 to 9.18 for the BNs studied.

## 1  Introduction

Bayesian networks (BNs) are an effective tool in a diverse range of applications that require representation and reasoning with uncertain knowledge and data. Inference over BNs can be either exact or approximate. Perhaps the most popular exact inference algorithm, belief propagation in junction trees, relies on the compilation of a BN into a junction tree. Exact belief updating (or marginalization) is then performed by message passing over the junction tree [6]. Each node of a junction tree is a clique computed from the moralized graph based on the original BN.

However, belief propagation over junction trees is known to be computationally hard. Computational difficulty increases dramatically with the density of the BN, the treewidth of the network, and the number of states of each network node [9]. In addition, some practical issues associated with the specific implementation platform also affect the computation performance [1]. In our work, we address the computational problem of belief propagation on both the analytical and implementation levels.

Two fundamental issues, which may cause large cliques in junction trees, are: (i) the topology and connectedness of a BN [9] and (ii) the high cardinality of a significant set of discrete BN nodes [13]. Discrete BN nodes can have high cardinalities for several reasons: First, they may represent discrete parameters, for example categorical parameters, that inherently take a large number of values [13]. A second reason for high-cardinality, discrete BN nodes is that they are used to represent continuous parameters. The number of states grows exponentially with the number of bits used when representing a quantized continuous parameter. Consequently, if a fine-grained discretization is used in BN nodes, the difficulty of computation may become a major challenge.

The above issues may cause very large cliques to be formed in junction trees, and thus hinder the application of BNs in cases where real-time inference is required. In addition, there can be major computational challenges when BN inference is in the inner loop of iterative algorithms like the EM algorithm [5]. Therefore, it is of great interest to develop parallel computing techniques to speed up junction tree inference. Recently, graphic processing units (GPUs) have become increasingly programmable and their parallel processing power can now be used for general purpose computation with Compute Unified Device Architecture (CUDA). However, due to the intricate nature of join tree computation and the distinctive GPU programming architecture, it is still a major challenge to adapt junction tree algorithms to the GPU. In this paper, we discuss data structures and algorithms that extend existing junction tree techniques [1,6], and specifically develop a novel approach to parallel message computation using belief propagation in junction trees.

[*]Email:lu.zheng, ole.mengsheol@sv.cmu.edu
[†]Email:jike@berkeley.edu

Parallelization of Bayesian network computation has been investigated in previous research [2–4, 7, 8, 10, 14, 15]. A data parallel implementation for junction tree inference has been developed for a cache-coherent shared-address-space machine with physically distributed main memory [4]. Parallelism in the basic sum-product computation has been investigated for GPUs [14]. The efficiency in using disk memory for exact inference, using parallelism and other techniques, has been improved [3]; parallel techniques for BN structure learning have also been developed [7]. An algorithm for parallel BN inference using pointer jumping has been introduced [10]. Both parallelization based on graph structure [8] as well as node level primitives for parallel computing based on a table extension idea have been developed [15]; a GPU implementation based on this idea was later developed [2].

In this paper, we also focus on node level parallelism, motivated by the existence of very large cliques in junction trees from applications. In such settings, node-level operations are often the dominating part of the problem [15]. However, we take a different approach from previous research [2, 15], and in particular our approach is motivated by the cluster-sepset mapping method of Huang and Darwiche [1]. We develop a parallel message computation algorithm for junction tree belief propagation. The speedup of this parallel algorithm, relative to the sequential algorithm, is analyzed theoretically. Experimental results, with speedups ranging from 0.68 to 9.18, show our GPU implementation's performance as it varies according to the junction tree topology.

Our paper is organized as follows: In Section 2, we review BNs, junction trees, and parallel computing using GPUs. In Section 3, we describe our parallel approach to message computation for belief propagation in junction trees. Experimental results are discussed in Section 4. In Section 5 we conclude and outline future research.

## 2 Background

### 2.1 Belief Propagation in Junction Trees

A BN is a compact representation of a joint distribution over a set of random variables $\mathcal{X}$. A BN is structured as a directed acyclic graph (DAG) whose vertices are the random variables and the directed edges represent dependency relationship among the random variables. The evidence in a Bayesian network consists of variables that have been instantiated.

The junction tree algorithm propagates beliefs (or posteriors) over a derived graph called a junction tree. A junction tree is generated from a BN by means of moralization and triangulation [6]. Each vertex $C_i$ of the junction tree contains a subset of the random variables that forms a clique in the moralized and triangulated BN, denoted by $\mathcal{X}_i \subseteq \mathcal{X}$. Associated with each vertex of the junction tree there is a potential table $\phi_{\mathcal{X}_i}$. With the above notations, a junction tree can be defined as $J = (\mathbb{T}, \Phi)$, where $\mathbb{T}$ represents a tree and $\Phi$ represents all the potential tables associated with this tree. Assuming $C_i$ and $C_j$ are adjacent, a separator $S_{ij}$ is induced on a connecting edge. The variables contained in $S_{ij}$ are defined to be $\mathcal{X}_i \cap \mathcal{X}_j$.

Belief propagation is invoked when we get new evidence $\boldsymbol{e}$ for a set of variables $\mathcal{E} \subseteq \mathcal{X}$. We need to update the potential tables $\Phi$ to reflect this new information. To do this, belief propagation over the junction tree is used, this is a two-phase procedure: evidence collection and evidence distribution. For the evidence collection phase, messages are collected from the leaf vertices all the way up to a designated root vertex. For the evidence distribution phase, messages are distributed from the root vertex to the leaf vertices.

Figure 1 shows a toy BN and the corresponding junction tree. As shown in the figure, during the evidence collection phase, the two leaf nodes pass messages to the root node $\{A, B, D\}$, updating its potential table. During the evidence distribution phase, the root node passes messages back to the two leaf nodes.

Message passing can be viewed as the atomic operation for belief propagation, both for evidence collection and distribution. Mathematically, a message passed from vertex $i$ to vertex $j$ can be written as:

$$\phi^*_{\mathcal{S}_{ij}} = \sum_{\mathcal{X}_i / \mathcal{S}_{ij}} \phi_{\mathcal{X}_i}, \quad \phi^*_{\mathcal{X}_j} = \phi_{\mathcal{X}_j} \frac{\phi^*_{\mathcal{S}_{ij}}}{\phi_{\mathcal{S}_{ij}}}, \qquad (1)$$

where $\phi^*_{\mathcal{X}}$ represents the updated potential table of vertex $\mathcal{X}$. From these potential tables we can marginalize and compute $P(X|\boldsymbol{e})$ where $X \in \mathcal{X}$ and $\boldsymbol{e}$ is the evidence.

### 2.2 CPU/GPU Platform

GPUs are designed for compute-intensive, highly parallel computations. Compared to CPUs, more transistors are in GPUs devoted to data processing rather than data caching and flow control. GPUs are well-suited to problems that can be expressed as data-parallel computations where data elements are mapped to parallel processing threads. GPUs are typically used to accelerate compute-intensive parts of an application, and thus attached to a host CPU performing control-dominant computations. Today, a CPU and its GPU communicate via a PCI-Express bus.

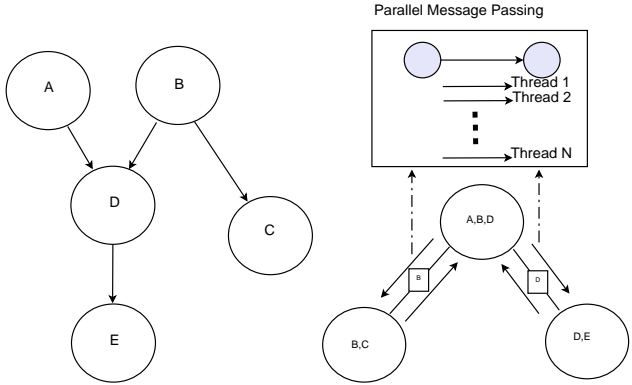

Figure 1: An example Bayesian network with five random variables (left) along with the derivative junction tree (right). Message passing is done between adjacent junction tree vertices, and we parallelize it using $N$ threads as indicated.

CUDA is a general-purpose parallel computing architecture developed by NVIDIA. CUDA consists of three key parallel computing abstractions: a hierarchy of thread groups, shared memories, and barrier synchronization. These abstractions are exposed to the programmer as a programming language, making CUDA a model that scales to an increasing number of processor cores [11]. Specifically, CUDA provides a fine-grained data and thread parallelism nested within coarse-grained data and task parallelism. A kernel is organized as a set of thread blocks when executed. A thread block is a batch of threads that all execute on one of the multiprocessors. As a result, they can cooperate by efficiently sharing data through shared memory, and can synchronize their execution to coordinate memory access. Blocks of the same size can execute the same kernel batched together as a grid of blocks. However, threads in different blocks cannot communicate and synchronize with each other.

## 3 Computing Each Message Faster

We parallelize the atomic operation of belief propagation–message passing, as shown in Figure 1. The advantage of doing so is that atomic level parallelism can be embedded in different belief propagation algorithms unobtrusively, without any change of those algorithms.

Associated with each junction tree vertex $\mathcal{C}_i$ and the contained set of variables $\mathcal{X}_i$, there is a potential table $\phi_{\mathcal{X}_i}$ containing non-negative real numbers that are proportional to the joint distribution of $\mathcal{X}_i$. If each variable can take $s_j$ states, the size of the potential table is $|\phi_{\mathcal{X}_i}| = \prod_{j=1}^{|\mathcal{X}_i|} s_j$, where $|\mathcal{X}_i|$ is the cardinality of $\mathcal{X}_i$.

Message passing from $\mathcal{C}_i$ to an adjacent vertex $\mathcal{C}_k$, with separator $\mathcal{S}_{ik}$, involves two steps:

1. **Marginalization**. The potential table $\phi_{\mathcal{S}_{ik}}$ of the separator is updated to $\phi^*_{\mathcal{S}_{ik}}$ by marginalizing the potential table $\phi_{\mathcal{X}_i}$:

$$\phi^*_{\mathcal{S}_{ik}} = \sum_{\mathcal{X}_i/\mathcal{S}_{ik}} \phi_{\mathcal{X}_i}. \qquad (2)$$

2. **Scattering**. The potential table of $\mathcal{C}_k$ is updated using both the old and new table of $\mathcal{S}_{ik}$:

$$\phi^*_{\mathcal{X}_k} = \phi_{\mathcal{X}_k} \frac{\phi^*_{\mathcal{S}_{ik}}}{\phi_{\mathcal{S}_{ik}}}. \qquad (3)$$

We define $\frac{0}{0} = 0$ in this case, that is, if the denominator in (3) is zero, then we simply set the corresponding $\phi^*_{\mathcal{X}_k}$ to zeros.

### 3.1 Index Mapping for Parallelism

Although written in a compact form, each of equation (2) and (3) is actually a set of many equations updating all the cells in the potential tables $\phi_{\mathcal{S}_{ik}}$ and $\phi_{\mathcal{X}_k}$. Our key contribution is to efficiently parallelize the computations in (2) and (3) by partitioning these sets of equations into independent subsets of equations.

This can be done by taking a closer look at the data flow in the message passing procedure. We concentrate on the $j$-th element of the separator's potential table, i.e., $\phi_{\mathcal{S}_{ik}}(j)$. Here, $\phi_{\mathcal{S}_{ik}}(j)$ is the potential value associated with a specific instantiation of the variables in $\mathcal{S}_{ik}$. In the marginalization step, to update the value of $\phi_{\mathcal{S}_{ik}}(j)$, we need to retrieve values from the elements in $\phi_{\mathcal{X}_i}$ which have the same instantiation for those variables. Note, the values of those elements in $\phi_{\mathcal{X}_i}$ are only required for the computation related to $\phi_{\mathcal{S}_{ik}}(j)$. Similarly, in the scattering step, for all the elements in $\phi_{\mathcal{X}_k}$ that have the same instantiation for the variables in $\mathcal{S}_{ik}$ as $\phi_{\mathcal{S}_{ik}}(j)$, we need to multiply their value by $\phi^*_{\mathcal{S}_{ik}}(j)/\phi_{\mathcal{S}_{ik}}(j)$. This suggests a natural way to compute (2) and (3) using data parallelism. To handle the computation related to each specific element in $\phi_{\mathcal{S}_{ik}}$, we can assign a separate thread, as long as the GPU has threads available.

Figure 2 illustrates the data flow in a message passing from the left child to the root node in Figure 1. In this case, $\mathcal{X}_i = \{A, B, D\}$, $\mathcal{X}_k = \{B, C\}$ and $\mathcal{S}_{ik} = \{B\}$. Let us assume that all random variables in $\{A, B, C\}$ are binary with states $\{0, 1\}$. The light gray boxes in Figure 2 are the values used in computations related to the first element of $\phi_{\mathcal{S}_{ik}}$, $B = 0$. The dark gray boxes are for computations related to the second element of $\phi_{\mathcal{S}_{ik}}$, $B = 1$. The computation related to different elements in $\phi_{\mathcal{S}_{ik}}$ are independent, providing a natural opportunity for parallelism.

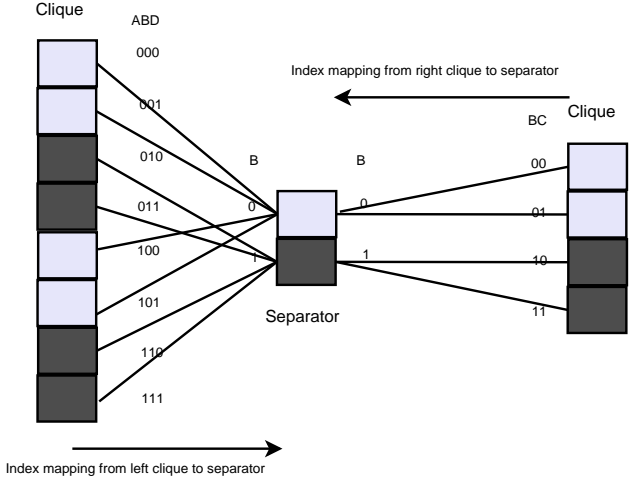

Figure 2: We introduce multiple index mapping tables between a clique and a separator. The mapping tables from the left clique to the separator: $\mu_{\mathcal{X}_i, \phi_{\mathcal{S}_{ik}}(0)} = \{0, 1, 4, 5\}$, $\mu_{\mathcal{X}_i, \phi_{\mathcal{S}_{ik}}(1)} = \{2, 3, 6, 7\}$. The mapping tables from the right clique to the separator: $\mu_{\mathcal{X}_k, \phi_{\mathcal{S}_{ik}}(0)} = \{0, 1\}$, $\mu_{\mathcal{X}_k, \phi_{\mathcal{S}_{ik}}(1)} = \{2, 3\}$.

The only problem left is that we need to figure out the "mapping relationship" from the elements of $\phi_{\mathcal{X}_i}$ and $\phi_{\mathcal{X}_k}$ to the elements of $\phi_{\mathcal{S}_{ik}}$. The mapping rule is that the corresponding elements in $\phi_{\mathcal{X}_i}$ and $\phi_{\mathcal{S}_{ik}}$ should have the same instantiation for the variables in $\mathcal{S}_{ik}$. To this end, for a certain element in $\phi_{\mathcal{S}_{ik}}$, say, $\phi_{\mathcal{S}_{ik}}(r)$, we first convert the index $r$ into a state string $Y^r = (x_1^r, \ldots, x_{|\mathcal{S}_{ik}|}^r)$ and then scan through all the elements of $\phi_{\mathcal{X}_i}$ or $\phi_{\mathcal{X}_k}$. For the $j$-th element in $\phi_{\mathcal{X}_i}$ ($j = 1, \ldots, |\phi_{\mathcal{X}_i}|$), we convert the index $j$ into a state string as $X_i^j = (x_1^j, \ldots, x_{|\mathcal{X}_i|}^j)$ and check whether the variables that also appear in $\mathcal{S}_{ik}$ take the same states as in $Y^r$. If yes, $\phi_{\mathcal{X}_i}(j)$ should be among the data to load in when we perform the marginalization for $\phi_{\mathcal{S}_{ik}}(r)$. Similarly, we can determine which elements should be updated in $\phi_{\mathcal{X}_k}$ during the scattering phase.

### 3.2 Index Mapping Table

Suppose the $j$-th random variable in $\mathcal{X}_i$ can take $s_j$ states. To convert the index $j$ into the sequence of variable states takes $O(|\mathcal{X}_i| \sum_j s_j)$. Then the whole scanning process to match the elements of $\mathcal{X}_i$ and $\mathcal{S}$ will take $O(|\phi_{\mathcal{X}_i}||\mathcal{X}_i| \sum_j s_j)$ time. This could be a considerable amount of computation time when the potential table size is large. It is extremely inefficient for every thread to scan through the whole potential table $\phi_{\mathcal{X}_i}$ and $\phi_{\mathcal{X}_k}$, since only a small fraction of them will be used by each thread. To tackle this potential inefficiency, we introduce an *index mapping table* technique inspired by the cluster-sepset mapping (CSM) technique [1], where a mapping table $\mu_{\mathcal{X}, \mathcal{S}}$ is created to store the index mappings from $\phi_{\mathcal{X}}$ to $\phi_{\mathcal{S}}$. To adapt CSM to parallel computing, instead of creating one mapping table [1], we create $|\phi_{\mathcal{S}_{ik}}|$ mapping tables. In each mapping table $\mu_{\mathcal{X}_i, \phi_{\mathcal{S}_{ik}}(j)}$ we store the indices of the elements of $\phi_{\mathcal{X}_i}$ mapping to the $j$-th separator table element. Mathematically, $\mu_{\mathcal{X}_i, \phi_{\mathcal{S}_{ik}}(j)} = \{r \in [0, |\phi_{\mathcal{X}_i}| - 1] : \phi_{\mathcal{X}_i}(r) \text{ is mapped to } \phi_{\mathcal{S}_{ik}}(j)\}$.

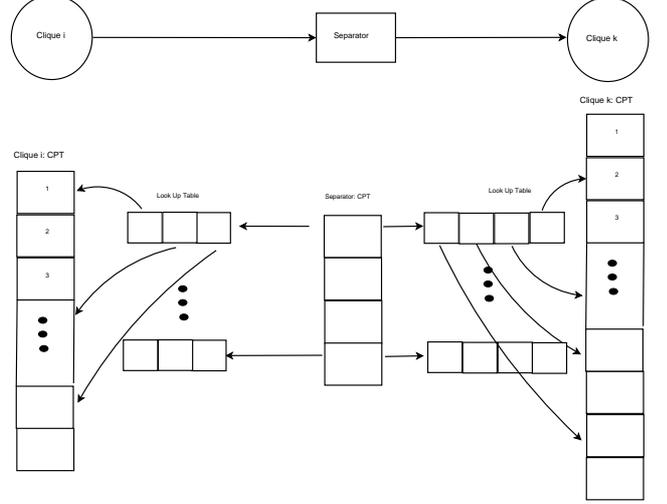

Figure 3: Data structure used to parallelize message passing from clique $m$ to clique $n$ in the junction tree.

---
**Algorithm 1** $Message\_Passing(\phi_{\mathcal{X}_i}, \phi_{\mathcal{X}_k}, \phi_{\mathcal{S}_{ik}})$
---
**Input:** $\phi_{\mathcal{X}_i}, \phi_{\mathcal{X}_k}, \phi_{\mathcal{S}_{ik}}$.
**for** $j = 1$ to $|\phi_{\mathcal{S}_{ik}}|$ **in parallel do**
  sep_star=0;
  **for** $n = 1$ to $|\mu_{\mathcal{X}_i, s_j}|$ **do**
    sep_star[j] = sep_star[j]+$\phi_{\mathcal{X}_i}(\mu_{\mathcal{X}_i, s_j}[n])$
  **end for**
  **for** $n = 1$ to $|\mu_{\mathcal{X}_k, s_j}|$ **do**
    $\phi_{\mathcal{X}_k}(\mu_{\mathcal{X}_k, s_j}[n]) = \frac{\text{sep\_star}[j]}{\phi_{\mathcal{S}_{ik}}[j]} \phi_{\mathcal{X}_k}(\mu_{\mathcal{X}_k, s_j}[n])$
  **end for**
**end for**

---

The thread that handles the $j$-th element of the separator potential table just needs to look up $\mu_{\mathcal{X}_i, \phi_{\mathcal{S}_{ik}}(j)}$ and retrieve the corresponding data from $\phi_{\mathcal{X}_i}$, as shown in Figure 3. Further, we avoid unnecessary recomputation of the mappings by precomputing them when the junction tree is established. Despite the requirement for memory (increased by the size of the clique potential table), our index mapping table often provides a large speedup.

Our novel algorithm for one message passing is shown in Algorithm 1. A function that runs on the GPU as different threads is called a *kernel*. Algorithm 1 is wrapped into a kernel function, thus enabling parallelism. While beneficial from a parallelism perspective, there are kernel invocation overhead and memory latency issues associated with this use of a GPU, as we will further discuss below.

### 3.3 Belief Propagation Algorithm

---
**Algorithm 2** Collect_Evidence($J, C_i$)
---
**for** each child of $C_i$ **do**
    Message_Passing($C_i$, Collect_Evidence($J$,child))
**end for**
return($C_i$)

---
**Algorithm 3** Distribute_Evidence($J, C_i$)
---
**for** each child of $C_i$ **do**
    Message_Passing($C_i$, child)
    Distribute_Evidence($J$, child)
**end for**

---
**Algorithm 4** Belief_Propagation($J, C_{root}$)
---
**Input:** $J, C_{root}$
Initialization ($J$)
Collect_Evidence($J, C_{root}$)
Distribute_Evidence($J, C_{root}$)

Belief propagation can be done using both breadth-first and depth-first traversal over a junction tree. In our work, we consider the Hugin algorithm, which adopts depth-first belief propagation. Given an established junction tree $J$ with root vertex $C_{root}$, the pseudo code is shown in Algorithm 4. We first initialize the junction tree by multiplying together the Bayesian network potential tables (CPTs). Then, a two phase belief propagation is adopted [6]: collect evidence and then distribute evidence.

### 3.4 Analysis of Speedup

From the description above, one can see that the amount of parallelism is determined by the number of elements in the separators' potential table $|\phi_S|$. Suppose the junction tree has $n$ vertices; then the total number of message passings for full belief propagation is $2(n-1)$. Considering a message passed from $C_i$ to $C_k$, the total number of additions is $(|\phi_{X_i}| - |\phi_{S_{ik}}|)$ and the total number of multiplications is $(|\phi_{X_k}| + |\phi_{S_{ik}}|)$. Therefore the theoretical time complexity of one message passing between vertex $i$ and $k$ is

$$\frac{(|\phi_{X_i}| - |\phi_{S_{ik}}|) + (|\phi_{X_k}| + |\phi_{S_{ik}}|)}{|\phi_{S_{ik}}|} = \frac{|\phi_{X_i}| + |\phi_{X_k}|}{|\phi_{S_{ik}}|},$$

which gives $|\phi_{S_{ik}}|$ times speedup over sequential code. Belief propagation is just a sequence of messages passed in a certain order [6].

Let $Ne(C)$ denote the neighbors of $C$ in the join tree. The time complexity for belief propagation is

$$\sum_i \sum_{k \in Ne(C_i)} \frac{|\phi_{X_i}| + |\phi_{X_k}|}{|\phi_{S_{ik}}|}. \qquad (4)$$

Kernel invocation overhead, incurred each time Algorithm 1 is invoked, turns out to be an important performance factor. If we model the invocation overhead for each kernel call to be a constant $\tau$, then the time complexity becomes

$$\sum_i d_i \tau + \sum_i \sum_{k \in Ne(C_i)} \frac{|\phi_{X_i}| + |\phi_{X_k}|}{|\phi_{S_{ik}}|}, \qquad (5)$$

where $d_i$ is the degree of a node $C_i$. In a tree structure, $\sum d_i = 2(n-1)$. Thus the GPU time complexity is

$$2(n-1)\tau + \sum_i \sum_{k \in Ne(C_i)} \frac{|\phi_{X_i}| + |\phi_{X_k}|}{|\phi_{S_{ik}}|}. \qquad (6)$$

From this equation, we can see that junction tree topology impacts GPU performance in at least two ways: the total invocation overhead is proportional to the number of nodes in the junction tree, while the separator table sizes determine the degree of parallelism.

The overall speedup of our novel parallel belief propagation approach is determined by the equation

$$Speedup = \frac{\sum_i \sum_{k \in Ne(C_i)} (|\phi_{X_i}| + |\phi_{X_k}|)}{2(n-1)\tau + \sum_i \sum_{k \in Ne(C_i)} \frac{(|\phi_{X_i}| + |\phi_{X_k}|)}{|\phi_{S_{ik}}|}}.$$

Clearly, performance is closely related to the distribution of the size of the separators' and cliques' potential tables. A simple bound for the speedup is $\min_{i,k} |\phi_{S_{ik}}| \leq Speedup \leq \max_{i,k} |\phi_{S_{ik}}|$. For the kind of junction tree that has mostly large separators, our parallel algorithm is expected to perform very well. The worst case is that all the separators of the junction tree are small. However, even in this case, since $|\phi_S| \geq 2$, we are in theory guaranteed to have at least two times speedup over sequential code. However, taking into account that the CPU/GPU platform incurs invocation overhead and the long memory latency when loading data from slow device memory to fast shared memory, the theoretical speedup is hard to achieve in practice.

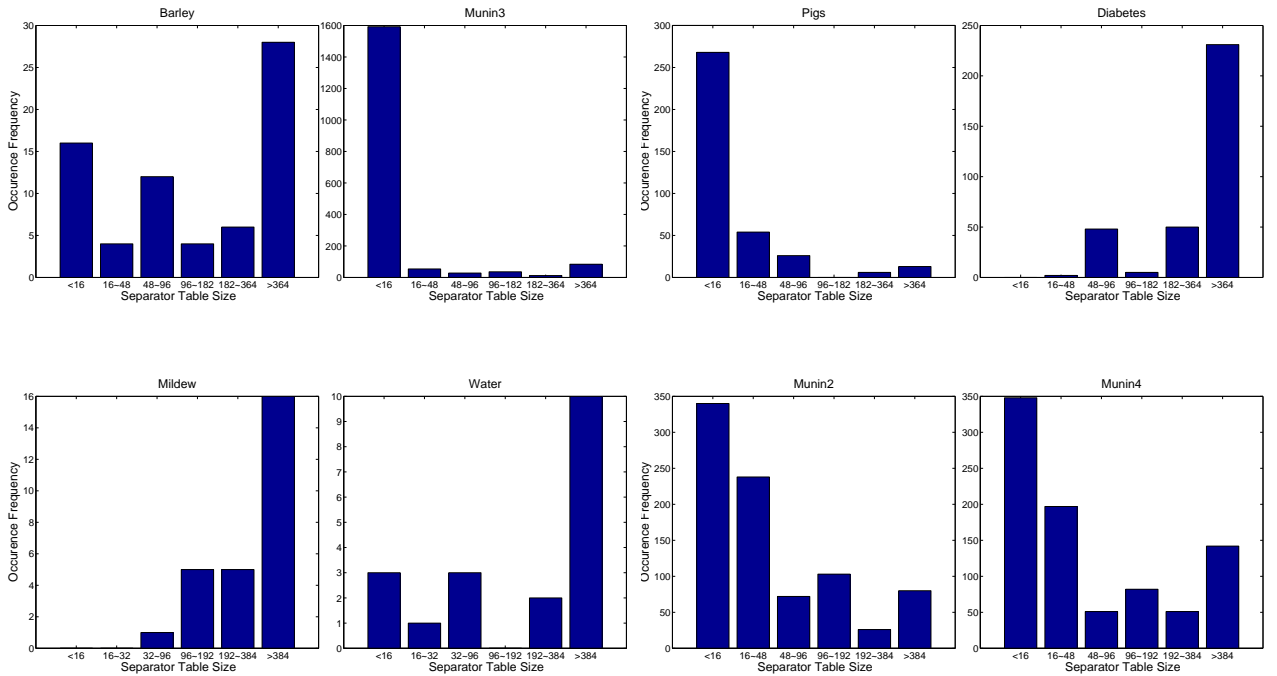

Figure 4: Histograms of separator potential table sizes for eight Bayesian networks.

From the equations above, we can estimate the overall belief propagation speedup to be around the average potential table size $|\phi^-_{\mathcal{S}_{ik}}|$. We take an experimental approach to study how the structure of the junction trees affects the performance of our parallel technique on the CPU/GPU setting in Section 4.

## 4 Experimental Results

In our work, we use the NVIDIA GeForce GTX460 as the platform for our implementation. This device consists of seven multiprocessors, and each multiprocessor consists of 48 cores and 48K on-chip shared memory per thread block. The peak thread level parallelism achieves 907GFlop/s. In addition to the fast shared memory, a much larger but slower off-chip global memory (785 MB) that is shared by all multiprocessors is provided. The bandwidth between the global and shared memories is about 90 Gbps. In the computation, we are using single precision.

### 4.1 Methods and Data

Our implementation is tested on a number of Bayesian networks (see http://bndg.cs.aau.dk/html/bayesian_networks.html). They are from different problem domains, with varying structures and state spaces. These differences lead to very different junction trees, as reflected in Table 1. In our work, we would like to not only to compare the performance of our parallel code to the sequential code, but also study how the structure of junction tree—for example the size of the separators' potential table—affects performance in the parallel case versus the sequential case. We compile the Bayesian networks into the junction trees offline and then run belief propagation over the junction trees, see Algorithm 4.

As mentioned in Section 3, the performance is related to the distribution of the size of the separators' potential table, i.e., $|\phi_{\mathcal{S}}|$. Hence we also present histograms of the potential table sizes for all the junction trees in Figure 4.

### 4.2 Optimization on GPU

Our novel message computation algorithm (Algorithm 1) is wrapped into a kernel to enable GPU parallelism, and a kernel is organized as a set of thread blocks when executed. Varying the thread block size may impact performance, and we would like to optimize thread block and grid size for each of the experimental junction trees. We experimented with varying block sizes and picked the best one for a given BN.

| Dataset | Mildew | Diabetes | Barley | Pigs | Munin2 | Munin3 | Munin4 | Water |
|---|---|---|---|---|---|---|---|---|
| # of JT nodes | 28 | 337 | 36 | 368 | 860 | 904 | 872 | 20 |
| Max. CPT size | 4,372,480 | 190,080 | 7,257,600 | 177,147 | 504,000 | 156,800 | 784,000 | 995,328 |
| Min. CPT size | 336 | 495 | 216 | 27 | 4 | 4 | 4 | 9 |
| Ave. CPT size | 341,651 | 32,443 | 512,044 | 1,927 | 5,653 | 3,443 | 16,444 | 173,297 |
| Max. SPT size | 71,680 | 11,880 | 907,200 | 59,049 | 72,000 | 22,400 | 112,000 | 147,456 |
| Min. SPT size | 72 | 16 | 7 | 3 | 2 | 2 | 2 | 3 |
| Ave. SPT size | 9,273 | 1,845 | 39,318 | 339 | 713 | 553 | 2,099 | 26,065 |
| BP on GPU [ms] | 53 | 94 | 106 | 75 | 125 | 104 | 342 | 52 |
| BP on CPU [ms] | 355 | 397 | 974 | 51 | 210 | 137 | 473 | 120 |
| Speedup | 6.70 | 4.22 | 9.19 | 0.68 | 1.68 | 1.32 | 1.38 | 2.31 |

Table 1: Junction tree (JT) statistics and belief propagation (BP) performance for eight Bayesian networks. For each junction tree, clique potential table (CPT) and separator potential table (SPT) are shown.

Figure 5: GPU execution time as a function of thread block size.

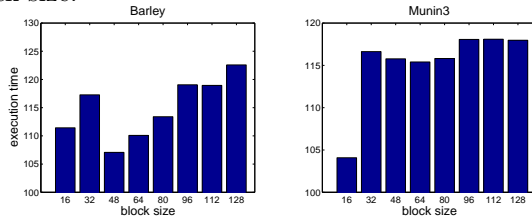

Figure 5 shows how execution time changes with block size for Bayesian networks *Barley* and *Munin3*. For *Barley*, GTX460 achieves the optimal performance when each block contains 48 threads. While for *Munin3*, optimal performance is found with 16 threads per block. The performance differences can be explained by the degree of match between the configuration of the GPU architecture and the junction tree structure. On the GPU, each thread block is executed on one multiprocessor. To fully make use of the GPU's computing resource, at least 7 thread blocks are needed, each assigned to one multiprocessor. However, for *Munin3*, the separator potential tables are very small. Less than 7 blocks are created for the message passing, leaving some multiprocessor unused. The size of *Barley*'s separator potential tables are mostly large enough to fully use the computing resource. In Figure 4, we present the histogram counts of the potential table sizes of *Barley* and *Munin3*.

Figure 6 illustrates the scalability of the speedup. We order the junction trees according to the average size of the separator potential tables, and plot the speedup relative to the average separator potential table. In general, the junction tree with larger average separator potential table has better speedup. This coincides with our analysis (see Section 3.4) that larger separator potential tables provide more opportunity for parallelization and hence better performance.

Figure 6: Speedup as a function of average separator potential table size.

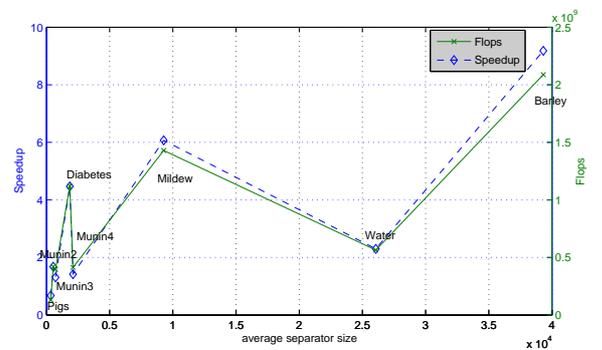

### 4.3 Performance Comparison with Sequential Code

As a baseline, we implemented a sequential program on an Intel CPU. The execution time of the program is comparable to that of GeNie/SMILE [12], a widely used C++ software package for Bayesian network inference. We do not directly use GeNie/SMILE as the baseline here, because we do not know the implementation details of GeNie/SMILE. Detailed information for the CPU and GPU platforms is in Table 2.

Table 1 gives the execution time comparison for the GTX460 and the Intel CPU. The obtained speedup ranges from 0.68 to 9.18. The performance is an overall effect of many factors such as parallelism, memory latency, kernel invocation overhead, etc. Those factors, in turn, are closely correlated with the underlying structures of the junction trees. Networks *Pigs*, *Munin2*, *Munin3* and *Munin4* mostly consist of small vertices and separators (see Figure 4). There are only limited opportunities for message computation parallelism, resulting in limited speedup. On the other hand, for *Mildew*, *Diabetes*, *Barley* and *Water*, the potential histograms in Figure 4 skew to the right. This explains why they have very good speedup. Two extreme examples are *Barley* and *Pigs*. The best performer, *Barley*, has a total of 36 cliques, and an average separator table size of 39,318. This topology provides abundant opportunity for parallelism and consequently a good speedup over the sequential code. *Pigs*, on the other hand, has 368 cliques, and furthermore the average separator size is as small as 339. This is a junction tree with a large number of small cliques and separators, with very restricted opportunity for parallelism in message computation.

Table 2: Experimental platforms.

| NVIDIA Geforce GTX 460 | |
|---|---|
| # of Processing Cores | 336 |
| Shared Memory | 48K per block |
| Global Memory | 785MB |
| Memory Bandwidth | 90 GB/sec peak |
| Intel Core2 Quad CPU | |
| # of Cores | 4 |
| Processor Clock | 2.5GHz |
| Cache | 8MB |
| Memory | 9 GB |

Figure 7: Comparison of kernel overhead (in yellow) and execution time (in green) for the datasets.

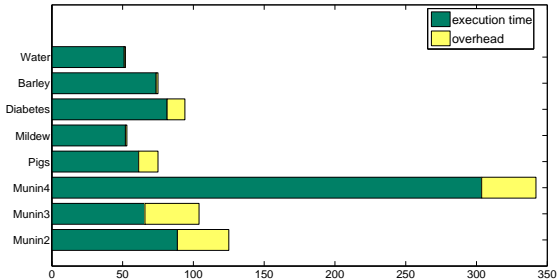

### 4.4 Kernel Overhead

For our parallel inference algorithm implementation on the GPU, we should also consider the overhead incurred when launching a kernel (kernel overhead). Figure 7 shows kernel overhead as a fraction of total execution time. Kernel overhead percentage is determined by the number of kernel invocations and the amount of computation per kernel invocation, which in turn is determined by the structure of the junction trees.

Kernel overhead may greatly affect the performance. For *Munin3*, for example, the overhead counts for as much as 36% in the overall execution time. This gives an upper bound of 2.71 on the speedup over sequential code. However, for *Barley*, the overhead is only 1.7% of the overall execution time. The variation in overhead percentages is caused by the differences in the structures of the *Munin3* and *Barley* junction trees, see Table 1 and Figure 4.

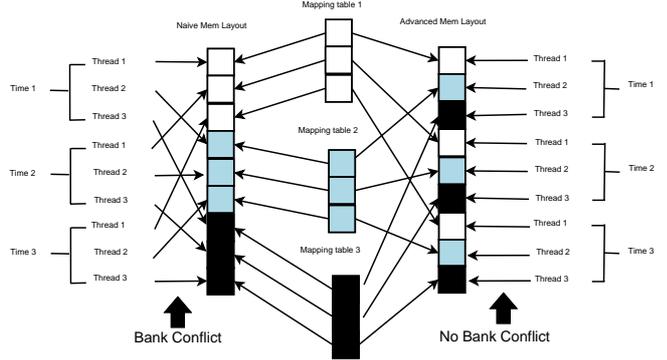

Figure 8: Memory layout for the mapping tables.

### 4.5 Memory Layout

Appropriate data layout in the global GPU memory makes a big difference in the memory latency associated with bringing data into shared GPU memory. In our algorithm, the mapping table method is essentially an indirect addressing for the data. In Figure 8, we compare two memory layouts for the mapping tables. On the left hand side of Figure 8 is a naive approach; mapping tables are just placed sequentially in global memory. This may cause bank conflicts when loading the data into the shared memory. Therefore, we introduce an advanced approach: for the mapping tables from a separator to a clique potential table, we put the elements with the same index in mapping tables in adjacent memory cells, as shown on the right hand side of Figure 8. In our experience, this advanced memory layout gave a 20% - 30 % improvement in the overall speedup compared to the naive layout.

## 5 Conclusion and Future Work

In this paper, we have developed a novel approach to parallel belief propagation over junction trees, based on the cluster-sepset mapping method. Our approach focuses on the parallelization of message computation for message passing in junction trees. In our approach, the parallel opportunity is in theory equal to the size of the separator potential table. Although practical issues such as kernel overhead and memory latency make it hard to achieve this theoretic performance, our experimental results still indicate that performance scales well with the separator potential table sizes.

In experiments with a CUDA implementation of our parallel message computation algorithm executing on an NVIDIA GeForce GTX460 GPU, we explored how performance varies with different junction tree structures. As expected from our analysis, we found that our approach performs well for junction trees with large separator potential tables. However, performance is compromised if the junction tree consists of many small nodes with corresponding small separator potential tables. Speedup ranged from 0.68 to 9.18.

Our future work will be focused on improving the parallel computing performance for junction tree message passing over small separators. A possible solution is merging small separators into large ones, and performing belief propagation over such modified junction trees.

**Acknowledgments**

This material is based upon work supported by NSF awards CCF0937044 and ECCS0931978.